\pdfoutput=1

\documentclass[11pt]{article}

\usepackage[final]{acl}

\usepackage{times}
\usepackage{latexsym}

\usepackage[T1]{fontenc}

\usepackage[utf8]{inputenc}

\usepackage{microtype}

\usepackage{inconsolata}

\usepackage{graphicx}
\usepackage{subcaption}
\usepackage{hyperref}
\usepackage{caption}
\usepackage{colortbl}
\usepackage{xspace}
\usepackage{enumitem}
\usepackage{booktabs}
\usepackage{multirow}
\usepackage{diagbox}
\usepackage{tabularx}
\usepackage[misc]{ifsym}
\usepackage{amsmath}
\usepackage{CJKutf8}

\let\oldfootnote\footnote
\renewcommand{\thefootnote}{\fnsymbol{footnote}}
\title{Learn Beyond The Answer: Training Language Models \\with Reflection for Mathematical Reasoning
}

\author{{\bf Zhihan Zhang\textsuperscript{\Letter}$^{1}$\thanks{\hspace{0.175cm}This work was done when Zhihan, Zhenwen, and Mengzhao were interns at Tencent AI Lab, Seattle.}, Tao Ge$^{2}$, Zhenwen Liang$^{1\dagger}$, Wenhao Yu$^{2}$},\\ {\bf Dian Yu$^{2}$, Mengzhao Jia$^{1\dagger}$, Dong Yu$^{2}$, Meng Jiang$^{1}$} \\
$^{1}$University of Notre Dame \hspace{0.2cm}
$^{2}$Tencent AI Lab, Seattle \hspace{0.2cm}
\\
\normalsize{ {\tt zzhang23@nd.edu}}
}

\begin{document}

\maketitle

\setcounter{footnote}{0}
\renewcommand{\thefootnote}{\arabic{footnote}}

\let\footnote\oldfootnote

\begin{abstract}
Supervised fine-tuning enhances the problem-solving abilities of language models across various mathematical reasoning tasks. To maximize such benefits, existing research focuses on \textit{broadening} the training set with various data augmentation techniques, which is effective for standard single-round question-answering settings. Our work introduces a novel technique aimed at cultivating a \textit{deeper} understanding of the training problems at hand, enhancing performance not only in standard settings but also in more complex scenarios that require reflective thinking. Specifically, we propose \textbf{reflective augmentation}, a method that embeds problem reflection into each training instance. It trains the model to consider alternative perspectives and engage with abstractions and analogies, thereby fostering a thorough comprehension through reflective reasoning. Extensive experiments validate the achievement of our aim, underscoring the unique advantages of our method and its complementary nature relative to existing augmentation techniques.\footnote{Code and data are available at \url{https://github.com/ytyz1307zzh/RefAug}.}

\end{abstract}

\section{Introduction}
\begin{figure*}[ht]
    \centering
    \includegraphics[width=\textwidth]{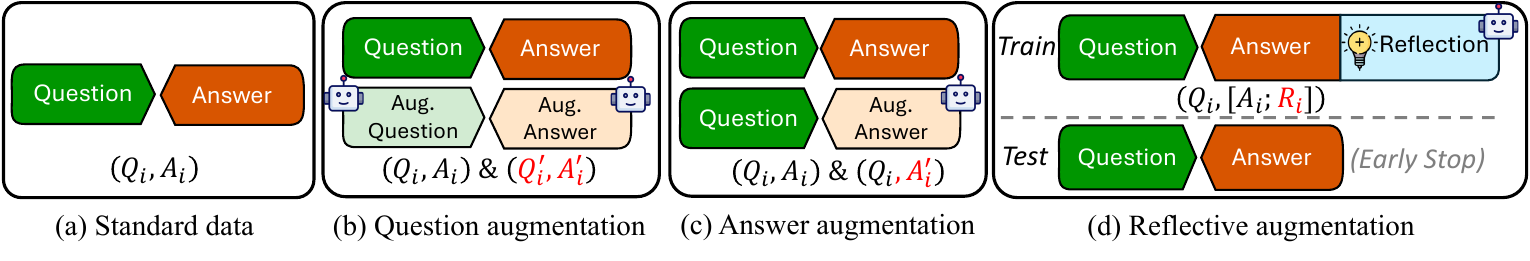}
    \vspace{-0.7cm}
    \caption{Question augmentation creates new questions based on existing ones. Answer augmentation re-samples answers for each problem to increase diversity. Both methods expand the size of the training set. Reflective augmentation appends the original answer with a \textbf{reflective section}, which is complementary to traditional approaches. Corresponding training sequences are shown in an (input, output) format, where augmented parts are in {\color{red}red}.}
    \label{fig:intro_illustration}
    \vspace{-0.1cm}
\end{figure*}

\begin{figure*}[t!]
    \centering

        \includegraphics[width=\textwidth]{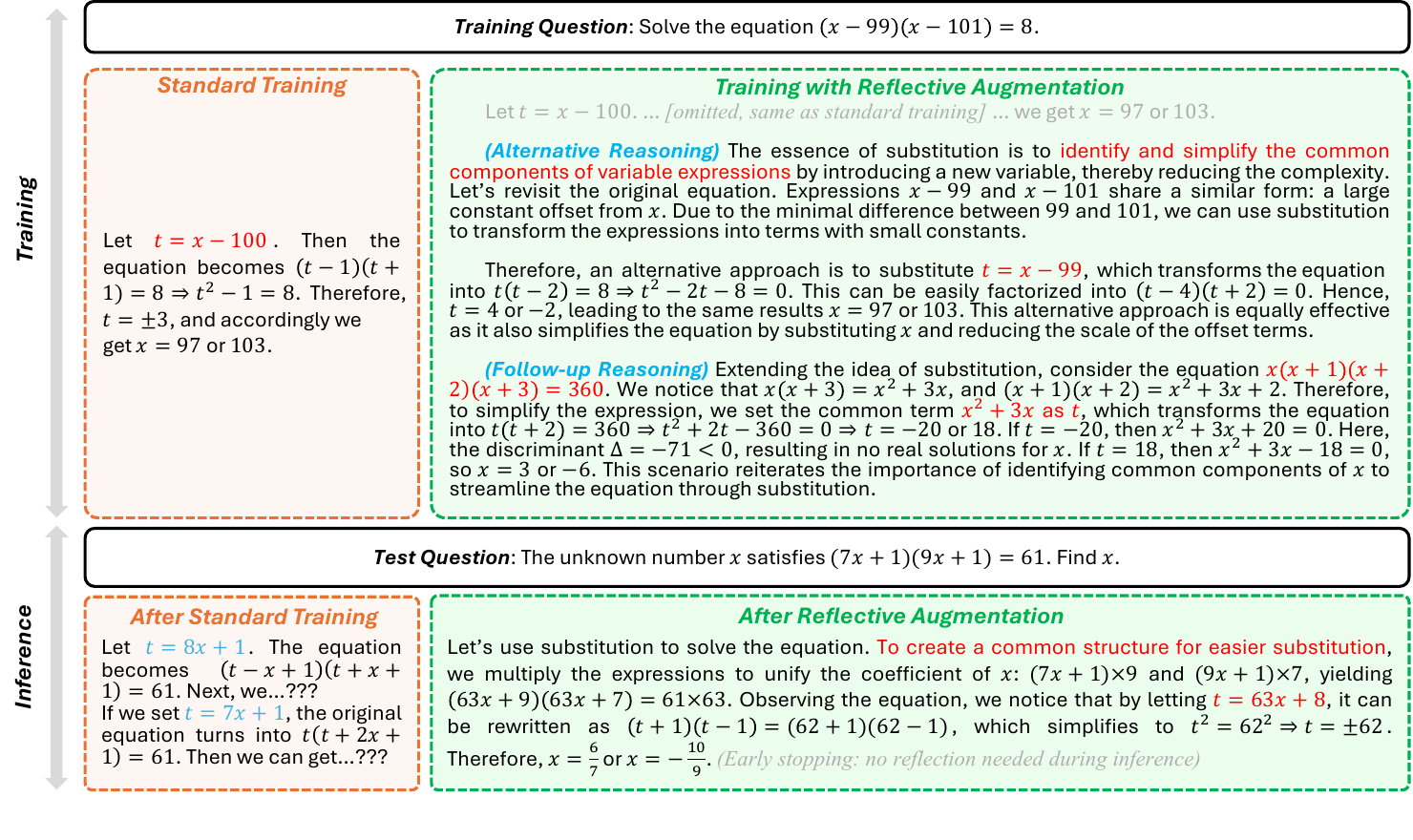}
        \vspace{-1cm}
        \caption{The model that learned the standard solution does not fully understand when and how to apply substitution when facing a different scenario. In contrast, the model trained with reflection on the substitution technique gains a deeper understanding of its principles, patterns, and its flexible application in new contexts.}
        \vspace{-0.2cm}
        \label{fig:intro_example}

\end{figure*}

The ability to engage in step-by-step reasoning is pivotal for language models (LMs) to solve mathematical problems \cite{chain-of-thought,zero-shot-cot}. Supervised fine-tuning, particularly on data with detailed reasoning paths, effectively advances the problem-solving performance of LMs \cite{small-lm-math-reason,mammoth}. To enlarge such benefits, most previous efforts focus on creating additional instances to augment model training \cite{wizardmath,metamath,orcamath,Xwin}. While these data expansion approaches allow LMs to handle a \textit{broader} range of math problems by increasing the diversity of training data, stacking more training instances does not necessarily lead to a \textit{deeper} understanding of each problem. Moreover, the scope of resulting models is confined to single-round question-answering (QA) settings that primarily require basic forward reasoning skills. Consequently, these methods provide limited benefits for more complex reflective reasoning scenarios that involve reviewing past steps for further reasoning, such as addressing follow-up questions, correcting errors, or leveraging external feedback~\cite{mathchat,mint}.

Similarly, the strategy in human learning is not always to practice an increasing number of problems~\cite{repetition-effect-human-learning}. 
Instead of merely memorizing superficial solutions to more problems, it can be more advantageous to gain a deep understanding of the existing problems~\cite{problem-based-learning}.
\textit{Reflection}, therefore, becomes an essential accompaniment to practice. \citet{think-mathematically} define reflection as ``to review thoughtfully, consider alternatives and follow extensions'', which encourages learners to contemplate their previous actions to engage in deeper reasoning, thereby fostering reflective thinking capabilities~\cite{alternative-1,followup-1}.

Inspired by such human cognition, we propose a novel training strategy for LMs that integrates reflection into each math problem. Unlike traditional data expansion methods which operate on the instance dimension by adding more training examples (see Figures~\ref{fig:intro_illustration}b \&~\ref{fig:intro_illustration}c), our approach targets a complementary direction, \textit{i.e.}, the sequence dimension of the training data. We introduce \textit{reflective augmentation} (\textbf{RefAug}), which appends a reflective section to the original answer of each training instance, advancing model learning beyond mere answer generation (see Figure~\ref{fig:intro_illustration}d). Such a design not only strengthens the model's understanding of the associated knowledge and methodologies in training problems, but also maintains the inference efficiency as the model ceases generation before decoding the reflective section during inference. Following the definition by \citet{think-mathematically}, these reflective sections include two components: \textit{alternative} and \textit{follow-up reasoning}. For example, Figure~\ref{fig:intro_example} shows a scenario where the model struggles to apply the substitution technique in a different context if only rigidly transferring the pattern from the standard solution. In contrast, training the model to reflect on an equivalent substitution expression followed by devising a more challenging equation facilitates a deeper understanding of the principles and variations of the technique, thereby enabling flexible adaptation in new contexts.


Extensive experimentation on diverse math reasoning tasks reveals multiple benefits of RefAug: (1) It boosts the problem-solving performance of LMs in the standard single-round QA settings, yielding a +7.2 accuracy gain over direct fine-tuning. (2) It remarkably enhances the LMs' performance in multiple reflective math reasoning scenarios, where traditional data expansion methods fall short. (3) Its benefits are complementary to those of existing data expansion techniques, allowing for seamless integration that leads to even greater performance improvements.

\section{Related Work}
\begin{figure*}[t]
    \centering
    \includegraphics[width=0.95\textwidth]{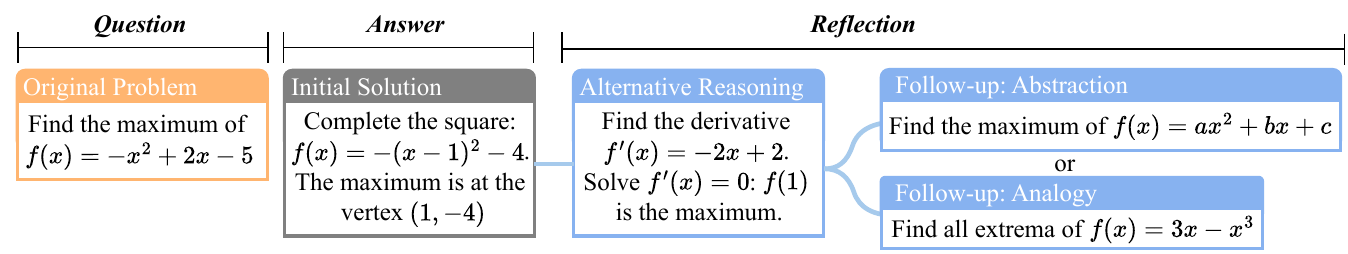}
    \vspace{-0.2cm}
    \caption{Relationship between the original instance and the reflective section. Either abstraction or analogy is annotated for each instance. Core ideas are shown but textual explanations (like those in Figure~\ref{fig:intro_example}) are omitted.}
    \label{fig:method}
    \vspace{-0.15cm}
\end{figure*}

\subsection{Data Augmentation for Math Reasoning}

Due to the scarcity~\cite{Xwin} and quality issues~\cite{reformat-alignment} of human-annotated data, data augmentation is a prevalent strategy in math reasoning tasks.
Most research focused on creating additional training instances, typically using advanced LMs to minimize human effort. This include \textit{question augmentation} which generates new questions from existing ones~\cite{metamath,mathscale,Xwin,MMIQC,KPMath}, and \textit{answer augmentation} which re-samples the answer for each question~\cite{RFT,mugglemath,metamath}. Others also explored \textit{answer refinement},  aiming to insert additional reasoning details~\cite{enrichmath} or to restructure answers for clearer reasoning paths~\cite{reformat-alignment}. Not only is reflective augmentation complementary to existing approaches, but it also exhibits unique advantages in reflective reasoning scenarios, as we will show in \S\ref{sec:experiments}.

Another branch of research augmented \textit{code snippets} within problem solutions, which transforms text reasoning into code generation~\cite{mathcoder,TORA,MathGenie}. This method is effective for math problems but is typically considered a separate track since it uses external tools (\textit{i.e.}, the code interpreter). Beyond supervised fine-tuning, some works augmented data for further preference optimization~\cite{iterative-reasoning-dpo,Eurus}, whereas we leave exploring reflective data in preference tuning for future work.

\vspace{-0.1cm}
\subsection{Reflection in LMs}
\label{sec:self_reflect}

Previous applications of reflection in LMs primarily focused on enabling LMs to rectify their own responses during inference (\textit{i.e.}, self-reflect). 
Some works equipped the LM with external feedback,  such as code execution or expert critiques~\cite{reflexion,self-debug}. Others prompted LMs to use only internal knowledge to correct answers~\cite{self-refine,testing-reflective-thinking}, though the effectiveness of this approach is under debate~\cite{cannot-self-correct}. Some specific tasks (\textit{e.g.}, math word problems) permit reverse verification, where the generated answer is used to re-derive the question to confirm its correctness~\cite{self-verification,Proco}. These works demonstrate that reflection is a common aspect of language processing. However, RefAug explores augmenting reflective data for better training instead of answer refinement during inference. Unifying these approaches is a promising future study.


\section{Approach}



RefAug extends each training sequence with a reflective section that encourages the LM to reflect on its initial reasoning process to engage in further math reasoning.
Figure~\ref{fig:intro_illustration} contrasts RefAug with traditional augmentation methods, and its detailed implementation is elaborated below.

\vspace{+0.1cm}
\noindent\textbf{Reflection Types} \quad Following the definition by \citet{think-mathematically} to ``review thoughtfully, consider alternatives and follow extensions'', we consider two types of reflection in composing the reflective section: \textit{alternative reasoning} and \textit{follow-up reasoning}.


Alternative reasoning involves thinking about the problem from different perspectives~\cite{alternative-1,alternative-2}. Therefore, besides the initial solution, we annotate an alternative approach that also effectively solves the problem. This helps the model master related methodologies and develop critical thinking skills.

Follow-up reasoning associates the initial solution to a broader class of problems~\cite{math-problem-posing,followup-question}. To fit various contexts, we consider two options: \textit{abstraction} and \textit{analogy}.
Abstraction refers to creating a generalized form of the original problem, thereby encouraging the model to reduce dependency on specific numerical values. Analogy challenges the model in applying methodologies of solving the original problem to a more complex situation. Learning to design follow-up scenarios enables the model to understand the associated math concepts and principles better and apply them flexibly in new contexts. The relationship between the initial instance and components of the reflective section is illustrated in Figure~\ref{fig:method}.

\vspace{+0.1cm}

\noindent\textbf{Data Annotation} \quad Following a common approach~\cite{mugglemath,metamath,Xwin}, we employ an expert LM, GPT-4-\textit{turbo}, to annotate the reflective sections for high-quality reasoning paths and minimal human effort\footnote{We also tried LLaMA-3-70B for data annotation in \S~\ref{sec:open-source-annotation} but its performance lags behind GPT-4-\textit{turbo}.}. This entails reviewing the original problem and solution to generate a section consisting of the aforementioned two types of reflective reasoning. We prompt the expert model to choose between abstraction and analogy in follow-up reasoning based on the problem context.
Figure~\ref{fig:intro_example} shows an annotated example with alternative reasoning and follow-up analogy, and the full annotation prompt is in Appendix~\ref{app:data_annotation_prompt}. The manual inspection and quality analysis of GPT-annotated data are detailed in Appendix~\ref{app:annotation_quality}.

\vspace{+0.1cm}
\noindent\textbf{Training \& Inference} \quad During training, given a math question as input,  we include the reflective section in the output immediately following the initial answer, starting with a \texttt{Reflection:} prefix. Thus, the training objective is to learn $\mathcal{P}([a;r]|q)$, where $[;]$ denotes sequence concatenation. Loss is calculated on tokens from both the initial answer and the reflective section. The format of the whole training sequence is detailed in Appendix~\ref{app:train_prompt}.

During inference, the generation early stops upon delivering the answer to the input question and ignores the reflective section, as shown in Figures~\ref{fig:intro_illustration}-\ref{fig:intro_example}. This is achieved by using \texttt{Reflection:} as a termination string during model generation.

\section{Experiments}
\label{sec:experiments}
We test RefAug in a variety of mathematical tasks that cover both standard single-round QA and reflective reasoning scenarios. We mainly evaluate two aspects: \textbf{the influence of RefAug on LMs' math reasoning abilities} and \textbf{its interaction with existing augmentation techniques}. Besides, we extend our approach to code generation tasks and perform comprehensive analyses.

\begin{table*}[t]
\centering
\resizebox{0.98\textwidth}{!}{
\begin{tabular}{c|c|cc|ccccc|c}
\toprule[0.15em]
\multirow{2}{*}{\textbf{Model}} & \multirow{2}{*}{\textbf{Training Data}} & \multicolumn{2}{c|}{\textit{In-Distribution}} & \multicolumn{5}{c|}{\textit{Out-Of-Distribution}}                                           & \multicolumn{1}{l}{\multirow{2}{*}{\textbf{Avg.}}} \\
&   & \textbf{GSM}               & \textbf{MATH}             & \textbf{Mathematics}    & \textbf{MAWPS}          & \textbf{SVAMP}          & \textbf{MMLU-Math}      & \textbf{SAT-Math}       & \multicolumn{1}{l}{}        
\\ \midrule[0.1em] 
\multicolumn{10}{c}{\textit{Closed-Source Models}}
\\
\midrule[0.1em] 
GPT-4-\textit{turbo} & -                       & 94.62             & 62.92            & 79.70          & 97.71          & 93.50          & 75.46          & 90.45          & 84.91 \\
GPT-3.5-\textit{turbo} & -                       & 74.68 & 44.36 & 64.70 & 94.27 & 82.40 & 61.70 & 77.27 & 71.34

\\ \midrule[0.1em] 
\multicolumn{10}{c}{\textit{Standard Training Data}}
\\
\midrule[0.1em]  
\multirow{2}{*}{Mistral} & Standard                       & 56.25             & 13.96            & 14.80          & 73.07          & 53.50          & 37.68          & 31.82          & 40.15                                     \\
& Standard + \textbf{RefAug}                  & \textbf{60.05}    & \textbf{17.36}   & \textbf{19.40} & \textbf{80.25} & \textbf{59.30} & \textbf{43.63} & \textbf{48.64} & \textbf{46.95}                            \\ \midrule[0.03em] 
\multirow{2}{*}{Gemma}     &    Standard    &   60.05    &   17.06    &   19.80    &   76.81    &   57.10    &   39.32    &   42.73    &   44.70 \\ 
&   Standard + \textbf{RefAug}     &   \textbf{64.59}    &   \textbf{23.04}    &   \textbf{26.70}    &   \textbf{85.64}    &   \textbf{64.70}    &   \textbf{46.61}    &   \textbf{55.00}    &   \textbf{52.33} \\   \midrule[0.1em] 
\multicolumn{10}{c}{\textit{Question Augmentation Data}}
\\
\midrule[0.1em]  
\multirow{3}{*}{Mistral}    &    Q-Aug                          & 56.03             & 18.06            & 18.00          & 79.99          & 59.10          & 38.19          & 36.16          & 43.65                                     \\
& Q-Aug$\times$2  & 59.14             & 21.26            & \textbf{20.90} & 80.84          & \textbf{61.50} & 40.86          & 46.82          & 47.33                                     \\
& Q-Aug + \textbf{RefAug}                     & \textbf{63.00}    & \textbf{21.66}   & 20.50          & \textbf{81.78} & 60.20          & \textbf{42.20} & \textbf{50.91} & \textbf{48.61}                            \\ \midrule[0.03em]
\multirow{3}{*}{Gemma}    &Q-Aug    &   61.11    &   21.98    &   23.90    &   81.78    &   59.70    &   40.45    &   48.18    &   48.16 \\ 
&   Q-Aug$\times$2    &   63.68    &   24.42    &   23.50    &   82.12    &   59.50    &   42.71    &   48.18    &   49.16 \\ 
&   Q-Aug + \textbf{RefAug}    &   \textbf{68.61}    &   \textbf{26.38}    &   \textbf{28.70}    &   \textbf{85.39}    &   \textbf{66.00}    &   \textbf{48.05}    &   \textbf{51.82}    &   \textbf{53.56} \\ \midrule[0.1em] 
\multicolumn{10}{c}{\textit{Answer Augmentation Data}}
\\
\midrule[0.1em]  
\multirow{4}{*}{Mistral} & A-Aug                          & 66.19             & 23.08            & 23.90          & 81.10          & 62.20          & 37.78          & 40.91          & 47.88                                     \\
& A-Aug$\times$2  & 67.93             & 27.12            & 28.30          & 83.26          & 66.50          & 42.61          & 45.91          & 51.66                                     \\
& A-Aug + Q-Aug                  & 69.67    & 24.32   & 26.90 & 81.82 & 61.20 & 38.50 & 46.82 & 49.90                            \\
& A-Aug + \textbf{RefAug}                     & \textbf{72.93}    & \textbf{29.40}   & \textbf{31.20} & \textbf{84.41} & \textbf{71.50} & \textbf{47.74} & \textbf{60.45} & \textbf{56.80}                            \\ \midrule[0.03em]

\multirow{3}{*}{Gemma}     &    A-Aug    &   68.31    &   28.78    &   33.10    &   83.05    &   65.10    &   46.51    &   61.36    &   55.17 \\ 
&   A-Aug$\times$2    &   70.66    &   31.14    &   33.30    &   85.22    &   \textbf{69.70}    &   47.13    &   54.55    &   55.96 \\ 
&   A-Aug + \textbf{RefAug}    &   \textbf{74.15}    &   \textbf{33.60 }   &   \textbf{38.20}    &  \textbf{ 85.68}    &   69.10    &   \textbf{52.26}    &   \textbf{64.09}    &   \textbf{59.58} \\ 
\midrule[0.1em] 
\multicolumn{10}{c}{\textit{MetaMath Augmentation Data}}
\\
\midrule[0.1em] 
\multirow{7}{*}{Mistral} & MetaMath\textsubscript{40k}                       & 68.46             & 20.96            & 20.30          & 85.09          & 66.50          & 38.09          & 42.73          & 48.88                                     \\
& MetaMath\textsubscript{80k}                       & 69.29             & 23.54            & 23.20          & 86.75          & 68.60          & 41.17          & 43.64          & 50.88                                     \\
& MetaMath\textsubscript{40k} + \textbf{RefAug}\textsubscript{40k}                  & \textbf{73.84}    & \textbf{26.60}   & \textbf{27.00} & \textbf{87.68} & \textbf{75.30} & \textbf{44.15} & \textbf{53.18} & \textbf{55.39}           \\ \cmidrule[0.03em]{2-10}     

& MetaMath\textsubscript{400k}* &  77.48 & 28.42 & 33.00 & 90.10 & \textbf{79.10} & 48.77 & 55.00 & 58.84 \\
& MetaMath\textsubscript{400k} + \textbf{RefAug}\textsubscript{40k} &  \textbf{78.70} & \textbf{32.50} & \textbf{34.50} & \textbf{91.59} & 77.90 & \textbf{49.69} & \textbf{59.09} & \textbf{60.57} \\

\cmidrule[0.03em]{2-10}     
& MetaMath\textsubscript{400k} (CT) &  78.39 & 28.72 & 32.70 & 90.87 & 78.90 & 49.08 & 55.91 & 59.22 \\

& MetaMath\textsubscript{400k} + \textbf{RefAug}\textsubscript{40k} (CT)  & \textbf{78.92} & \textbf{30.12} & \textbf{36.20} & \textbf{91.46} & \textbf{79.90} & \textbf{49.69} & \textbf{57.27} & \textbf{60.51}
\\ \bottomrule[0.15em]
\end{tabular}}
\vspace{-0.1cm}
\caption{Accuracy on single-round math reasoning tasks. * The public checkpoint released by~\citet{metamath}.}
\vspace{-0.2cm}
\label{tab:math-main}
\end{table*}

\subsection{Standard Math Reasoning}

\label{sec:standard_math}

\subsubsection{Settings}

Standard math reasoning tasks follow a single-round QA format. Following a popular approach, we use the training sets of GSM8k \cite{gsm8k} and MATH \cite{MATH}. We additionally include out-of-distribution test sets from  MAWPS \cite{MAWPS}, Mathematics \cite{mathematics}, SVAMP \cite{svamp}, plus the math subsets of MMLU \cite{MMLU} and SAT \cite{AGIEval}.
We mainly experiment with two LMs known for superior reasoning performance: Mistral-7B~\cite{mistral} and Gemma-7B~\cite{gemma}, and have also tested LLaMA-3-8B~\cite{llama-3} in Appendix~\ref{app:llama3}. Models are trained for 3 epochs with batch size 128. The learning rate peaks at 1e-5 with a 3\% warmup period followed by linear decay. Greedy decoding is applied during inference. Additional details of datasets and training settings are in Appendix~\ref{app:standard_math_task_settings}.

\subsubsection{Existing Training Methods}
\label{sec:baseline}


\begin{itemize}
[noitemsep,topsep=3pt,parsep=1pt,partopsep=0pt,leftmargin=*]

\item \textbf{Standard Fine-tuning} (Figure~\ref{fig:intro_illustration}a): Utilizes original problem solutions from GSM8k and MATH, each containing a chain-of-thought reasoning process before reaching the final prediction.
\item \textbf{Question Augmentation} (\textbf{Q-Aug}, Figure~\ref{fig:intro_illustration}b): Involves training on both original and GPT-augmented questions. We adopt the augmentation prompt from~\citet{Xwin}, detailed in Appendix~\ref{app:baseline}. We also explore \textbf{Q-Aug + RefAug} by applying RefAug to all questions after Q-Aug, and \textbf{Q-Aug$\times$2} by adding a second augmentation round to further expand the dataset.

\item \textbf{Answer Augmentation} (\textbf{A-Aug}, Figure~\ref{fig:intro_illustration}c): Re-samples the solution for each problem using GPT-4-\textit{turbo}, following the approach of~\citet{metamath}. We also explore its combination with Q-Aug (\textbf{A-Aug + Q-Aug}), RefAug (\textbf{A-Aug + RefAug}), and another round of A-Aug (\textbf{A-Aug$\times$2}).

\item \textbf{MetaMath Augmentation:} MetaMath~\cite{metamath} creates a training set of 400K instances using various augmentation techniques. Due to budget constraints, we examine the following subsets: (1) A uniformly sampled 40K subset (\textbf{MetaMath\textsubscript{40k}}), which we augment with RefAug to compare against an 80K sample (\textbf{MetaMath\textsubscript{80k}}); (2) The entire 400K dataset, of which 40K instances are augmented with RefAug (\textbf{MetaMath\textsubscript{400k}+RefAug\textsubscript{40k}}), to compete with the public MetaMath checkpoint; (3) A one-epoch continual training (\textbf{CT}) from the public checkpoint on the same dataset as (2).

\end{itemize}

The augmentation prompt for Q-Aug and A-Aug, along with the sampling strategy on MetaMath can be found in Appendix~\ref{app:baseline}.

\begin{table*}[t]
\centering

    \resizebox{0.87\textwidth}{!}{
\begin{tabular}{c|ccc|c|cccccc}
\toprule
\multirow{2}{*}{\textbf{Training Data}} & \multicolumn{3}{c|}{\textbf{MathChat-FQA}}               & \multirow{2}{*}{\textbf{MathChat-EC}}  & \multicolumn{6}{c}{\textbf{MINT-Math}} \\
                      & \textit{1st}            & \textit{2nd}            & \textit{3rd}            &      & $k=1$            & $k=2$            & $k=3$            & $k=4$            & $k=5$            & \textit{$\Delta$}                              \\ \midrule
Standard                  &     56.25              &     25.72              &     15.25              &     50.68         &    20.88    &    24.91    &    27.47    &    28.57    &    28.94    &    8.06              \\
Standard + \textbf{RefAug}             &     \textbf{60.05}     &     \textbf{35.36}     &     \textbf{27.54}     &     \textbf{72.99}          &   \textbf{22.34}   &    \textbf{33.70}    &    \textbf{37.00}    &    \textbf{38.10}    &    \textbf{39.56}    &    \textbf{17.22}    \\ \midrule
Q-Aug                     &     56.03              &     30.65              &     21.02              &     65.48            &    21.98    &    27.47    &    30.04    &    31.87    &    32.60    &    10.62           \\
Q-Aug$\times$2              &    59.14              &     32.70              &     22.99              &     63.51         &    \textbf{27.11}    &    32.60    &    35.16    &    36.26    &    37.73    &    10.62              \\
Q-Aug + \textbf{RefAug}                &     \textbf{63.00}     &     \textbf{42.19}     &     \textbf{34.37}     &     \textbf{76.48}       &    26.74    &    \textbf{37.36}    &    \textbf{41.03}    &    \textbf{42.86}    &    \textbf{43.22}    &    \textbf{16.48}       \\ \midrule
A-Aug                     &     66.19              &     34.29              &     23.60              &     72.08          &    23.08    &    30.77    &    33.70    &    35.16    &    35.53    &    12.45             \\
A-Aug$\times$2              &     67.93              &     36.57              &     28.00              &     71.93           &    25.64    &    31.87    &    33.33    &    34.80    &    34.80    &    9.16            \\
A-Aug + Q-Aug             &     69.67              &     37.86              &     27.31              &     69.58     & 23.44 & 31.87 & 35.16 & 37.36 & 38.10 & 14.66      \\
A-Aug + \textbf{RefAug}                &     \textbf{72.93}     &     \textbf{44.92}     &     \textbf{36.19}     &     \textbf{80.20}         &    \textbf{28.94}    &    \textbf{42.12}    &    \textbf{46.15}    &    \textbf{47.28}    &    \textbf{47.99}    &    \textbf{19.05}              \\ \midrule
MetaMath              & 68.46          & 37.48          & 24.89          & 61.15   & 22.34 & 27.84 &   31.50 & 32.23 & 33.70   & 11.36           \\
MetaMath$\times$2              & 69.29          & 38.92          & 26.10          & 60.09    &  21.61 & 25.64 & 26.74 &  27.47 & 27.84  & 6.23           \\
MetaMath + \textbf{RefAug}         & \textbf{73.84} & \textbf{43.93} & \textbf{34.98} & \textbf{79.51} & \textbf{27.47} &  \textbf{36.63} & \textbf{39.93} &  \textbf{40.66} & \textbf{41.03}  & \textbf{13.56}     \\ \bottomrule
\end{tabular}}
\vspace{-0.1cm}
\caption{Accuracy on reflective math reasoning tasks. Each question in MathChat-FQA has two subsequent questions (\textit{2nd} and \textit{3rd} turns), and the accuracy of each turn is calculated separately. MINT evaluates whether the model solves the math problem within $k$ interaction turns with the feedback from GPT-4, and we use the difference ($\Delta$) between $k=5$ and $k=1$ to indicate the model's ability in leveraging external feedback.}
\vspace{-0.2cm}
\label{tab:math-reflect}
\end{table*}

\subsubsection{Results}
\label{sec:standard-math-results}

Table~\ref{tab:math-main} lists the QA accuracy of fine-tuned LMs. We summarize several findings on RefAug:

\textbf{Enhancement in Single-Round Math Reasoning}: RefAug boosts model performance across both in-distribution and out-of-distribution tasks, outscoring the direct fine-tuning approach by +7.2 across two base LMs. As the reflective section is not utilized during inference, this advancement underscores RefAug's role in enhancing model learning, which strengthens math problem-solving capabilities without providing additional context.



\textbf{Complementary Benefits with Existing Methods}: While data expansion methods (Q-Aug, A-Aug, and MetaMath) have improved model performance, combining RefAug with them leads to further substantial gains, improving overall accuracy by +6.1 on average. This demonstrates that RefAug still holds value on high-quality data\footnote{\label{note:gpt_answer_quality}In Appendix~\ref{app:gpt-answer}, we show that GPT-written solutions are of higher quality than those original ones in GSM and MATH.} and is complementary to data expansion strategies. Furthermore, such synergistic benefits outpace the diminishing returns seen with repeated dataset expansions: these three methods bring +6.8 improvement initially but only +2.3 in the second round. This disparity indicates that expanding data does not always yield proportionate gains, whereas the balance of practicing new problems and reflecting on existing ones maximizes the learning effect.

\textbf{Effectiveness on Large Datasets}: Even when only 10\% of the full-sized MetaMath dataset includes the reflective section, the resulting model surpasses the public MetaMath checkpoint by \textasciitilde2 points. This confirms RefAug's efficacy on larger scales of data. Additionally, the MetaMath model barely benefits from continual training on its original QA data, suggesting a good memorization of these math problems. Nevertheless, RefAug still manages to elevate its performance, indicating that the model has not fully internalized the dataset's knowledge and RefAug effectively deepens the model's understanding of these problems.



\subsection{Reflective Math Reasoning}

\label{sec:reflect-math}

\subsubsection{Tasks}

Many realistic math applications require models to reflect on previous predictions and perform further reasoning. We employ three tasks of this kind: the follow-up QA (\textbf{FQA}) and error correction (\textbf{EC}) tasks of MathChat~\cite{mathchat}, and the math subset of \textbf{MINT}~\cite{mint}. FQA involves solving two subsequent questions linked to each initial query, forming a three-round interaction. EC deliberately writes an erroneous solution to test the model's error identification and correction abilities. MINT evaluates the model's ability to leverage external language feedback to improve its reasoning process through up to $k$ turns of interaction. More task details are in Appendix~\ref{app:reflective_math_task_settings}.

\begin{table}[t]
\centering
\setlength{\tabcolsep}{1.3mm}{
\resizebox{0.48\textwidth}{!}{
\begin{tabular}{l|c|ccc|c}
\toprule
\multirow{2}{*}{\textbf{Model}}                 & \multirow{2}{*}{\textbf{Data}}              & \multicolumn{2}{c}{\textbf{FQA}}               & \multirow{2}{*}{\textbf{EC}}  & \multirow{2}{*}{\textbf{Avg.}}           \\
& & \textit{2nd} & \textit{3rd} & & \\
\midrule
GPT-4-\textit{turbo}               & -              & 77.67          & 73.03          & 83.09  & 77.93        \\
GPT-3.5-\textit{turbo}              & -              & 55.26          & 45.59          & 75.90   & 58.92       \\

\midrule
MAmmoTH               & 184K              & 32.16          & 19.31          & 54.15  & 35.21        \\
MetaMath              & 395K              & 43.98          & 32.16          & 56.30   & 44.15       \\
WizardMath            & 112K*              & 44.81          & \underline{36.86} & 68.22   & 49.96       \\
InternLM2-Math        & \textasciitilde2M & 40.20          & 28.64          & 72.70   & 47.18       \\
DeepSeek-Math         & 776K              & \textbf{48.19} & 35.70          & 74.34    & 52.74 \\
Mistral+A-Aug+\textbf{RefAug} & 30K               & 44.92    & 36.19    & \underline{80.20} & \underline{53.77} \\
Gemma+A-Aug+\textbf{RefAug} & 30K & \underline{47.80} & \textbf{38.54} & \textbf{81.11} & \textbf{55.82}\\ \bottomrule
\end{tabular}}}
\caption{MathChat results compared with other open-source 7B math models. Baseline scores are from~\citet{mathchat}. The best scores are \textbf{bolded} and the second bests are \underline{underlined}. GPT models are listed as a reference for state-of-the-art performance. *Including both supervised fine-tuning and reinforcement learning data.}
\vspace{-0.3cm}
\label{tab:mathchat}
\end{table}

\subsubsection{Results}

Results on reflective math reasoning tasks are displayed in Tables~\ref{tab:math-reflect}-\ref{tab:mathchat} for Mistral and Table~\ref{tab:math-reflect-gemma} for Gemma. We summarize the key findings below.

\textbf{Challenges for Data Expansion Methods}: Despite improving single-round QA performance, methods like Q-Aug, A-Aug, and MetaMath fall short in enhancing LMs' reflective reasoning abilities. For instance, these methods hurt Mistral's error correction performance. Moreover, a second round of augmentation yields minimal or negative gains across key metrics on reflective reasoning: +2.5 in FQA-\textit{3rd}, -1.1 in EC, -0.5 in MINT$_{k=5}$, and -4.2 in MINT$_{\Delta}$. This indicates that initial augmentation benefits are mainly due to the improved answer quality from GPT annotation\textsuperscript{\ref{note:gpt_answer_quality}} rather than an actual increase in reflective reasoning skills, which echos the findings of~\citet{mathchat} that conventional training approaches overly focus on the single-round QA setting and neglect many other important mathematical scenarios.

\begin{table*}[t]
\centering
\setlength{\tabcolsep}{1.3mm}{
\resizebox{0.92\textwidth}{!}{
\begin{tabular}{l|cc|ccccc|c}
\toprule
\textbf{Data}          & \textbf{GSM}   & \textbf{MATH}  & \textbf{Mathematics} & \textbf{MAWPS} & \textbf{SVAMP} & \textbf{MMLU-Math} & \textbf{SAT-Math} & \textbf{Avg.}  \\ \midrule
Standard                 & 56.25          & 13.96          & 14.80                & 73.07          & 53.50          & 37.68              & 31.82             & 40.15          \\
\quad+ Alternative Reasoning    & 59.51          & 16.42          & 17.90                & 79.57          & 58.30          & 39.63              & 44.09             & 45.06          \\
\quad+ Follow-up Reasoning & 56.25          & 16.82          & 18.80                & 77.10          & 58.50          & 38.09              & 44.05             & 44.23          \\
\quad+ \textbf{RefAug}    & \textbf{60.05} & \textbf{17.36} & \textbf{19.40}       & \textbf{80.25} & \textbf{59.30} & \textbf{43.63}     & \textbf{48.64}    & \textbf{46.95} \\ \bottomrule
\end{tabular}}}
\vspace{-0.1cm}
\caption{Accuracy on standard math reasoning tasks when varying the components of the reflective section.}
\label{tab:ablation}
\vspace{-0.2cm}
\end{table*}

\textbf{Superiority of RefAug in Enhancing Reflective Reasoning}: RefAug significantly enhances the model's reflective reasoning performance, with gains of +12.3 in FQA-\textit{3rd}, +22.3 in EC, +10.6 in MINT$_{k=5}$, and +9.2 in MINT$_{\Delta}$, far exceeding the corresponding improvements of +7.9, +15.5, +5.0, and +3.4 brought by three data expansion methods on average. An effective solution, however, is to combine RefAug with these methods, which yields substantial improvements over them, \textit{e.g.}, +12 on FQA-\textit{3rd} and +10.1 on MINT$_{k=5}$. These results highlight RefAug's exceptional capability to improve LMs' reflective math reasoning, which complements the disregard of existing augmentation methods on this dimension.

\textbf{Comparison with Existing Open-Source Models}: Our RefAug-enhanced models excel in the reflective reasoning scenarios of MathChat with just 30K training instances, surpassing many open-source models trained on larger math datasets or with reinforcement learning. This further supports RefAug's effectiveness in cultivating LMs' reflective reasoning skills in solving math problems.


Based on findings from \S\ref{sec:standard_math} and \S\ref{sec:reflect-math}, we conclude the benefits of RefAug on math reasoning as:
\textbf{\textit{Not only does it enhance LMs' basic problem-solving skills but also advances their reflective reasoning abilities, making it a valuable complement to existing augmentation techniques.}}

\begin{table}[t]
\centering
    \setlength{\tabcolsep}{1.3mm}{
    \resizebox{0.48\textwidth}{!}{
\begin{tabular}{l|cccc|r}
\toprule
\textbf{Model}       & \textbf{HE} & \textbf{HE+} & \textbf{MBPP} & \textbf{MBPP+} & \multicolumn{1}{c}{\textbf{Avg.}} \\ \midrule
CodeLlama-std     & 53.7               & 50.6                & 62.9          & 51.6           & 54.7                             \\
CodeLlama-\textbf{RefAug}       & \textbf{57.9}      & \textbf{53.0}       & \textbf{65.4} & \textbf{52.4}  & \textbf{57.2}                    \\\midrule
Mistral-std       & 38.4               & 35.4                & 53.1          & 40.1           & 41.7                             \\
Mistral-\textbf{RefAug}         & \textbf{50.0}        & \textbf{45.1}       & \textbf{56.4} & \textbf{46.4}  & \textbf{49.5}                    \\\midrule
StarCoder2-std    & 54.3               & 49.4                & 62.7          & 51.4           & 54.4                             \\
StarCoder2-\textbf{RefAug}      & \textbf{56.7}      & \textbf{50.6}       & \textbf{66.7} & \textbf{51.6}  & \textbf{56.4}                    \\\midrule
DeepSeekCoder-std & \textbf{67.1}      & 59.8                & 75.4          & 60.4           & 65.7                             \\
DeepSeekCoder-\textbf{RefAug}   & \textbf{67.1}      & \textbf{62.2}       & \textbf{76.7} & \textbf{63.2}  & \textbf{67.3}                    \\ \bottomrule
\end{tabular}}}
\vspace{-0.1cm}
\caption{Pass@1 on code generation, scored by EvalPlus. -std denotes training with the standard QA setting.}
\vspace{-0.3cm}
\label{tab:code}
\end{table}

\subsection{Code Generation}

Besides math reasoning, we extend the application of RefAug to code generation. In this task, a query instructs the model to craft a code snippet that fulfills a specific functionality, which also requires a step-by-step logical flow. We use HumanEval \cite{humaneval} and MBPP \cite{MBPP} as the evaluation benchmarks, along with their plus versions provided by EvalPlus \cite{evalplus}. Training is conducted using the Python subset of Magicoder-OSS-Instruct \cite{magicoder}, which includes 38K QA instances. Considering the abstractive nature of code, we annotate problem analogies as the follow-up section of RefAug.

The outcomes are summarized in Table ~\ref{tab:code}, covering four different base LMs: CodeLLaMA~\cite{codellama}, Mistral, StarCoder2~\cite{starcoder2}, and DeepSeekCoder~\cite{deepseekcoder}. The results demonstrate that RefAug consistently elevates the LMs' proficiency in following instructions to generate accurate, reasonable code, as evidenced by an average improvement of +3.5 in Pass@1 across the evaluated benchmarks. These results indicate that RefAug is able to enhance LMs' capabilities in solving code problems, which reaffirms from another scenario that reflection is an essential ability for LMs to possess.


\begin{figure}[t]
    \centering
    \includegraphics[width=0.4\textwidth]{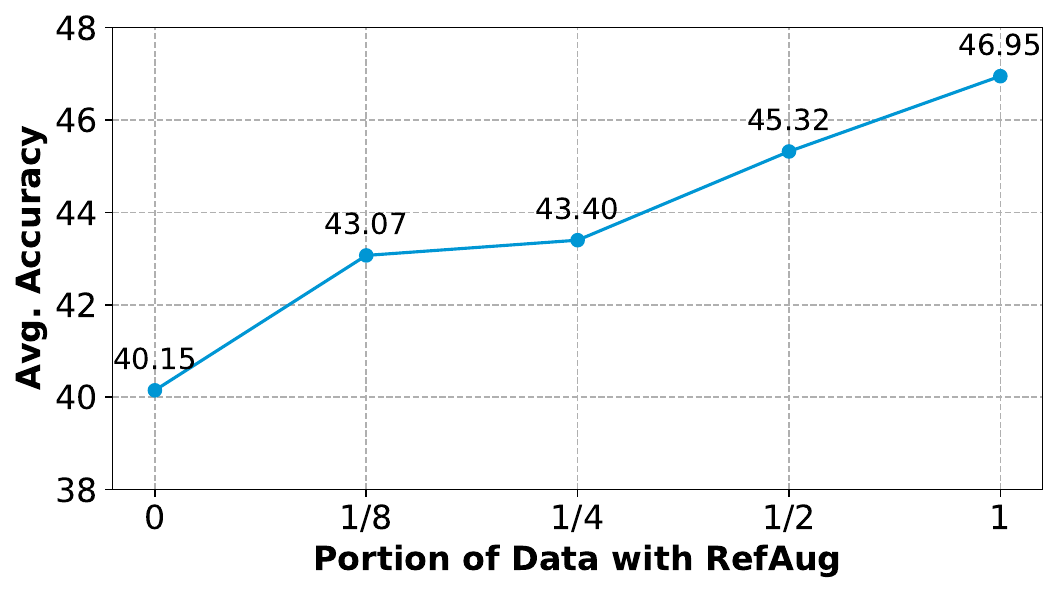}
    \vspace{-0.2cm}
    \caption{Average accuracy on 7 standard math reasoning tasks when different proportions of data are augmented with reflective sections (remaining data are in the standard QA form).}
    \vspace{-0.3cm}
    \label{fig:data_scaling}
\end{figure}


\begin{table*}[t]
\centering
\setlength{\tabcolsep}{1.3mm}{
\resizebox{0.95\textwidth}{!}{
\begin{tabular}{l|cc|ccccc|c|ccc}
\toprule
\textbf{Data}          & \textbf{GSM}   & \textbf{MATH}  & \textbf{Mathematics} & \textbf{MAWPS} & \textbf{SVAMP} & \textbf{MMLU} & \textbf{SAT} & \textbf{Avg.} & \textbf{FQA}-\textit{2nd} & \textbf{FQA}-\textit{3rd} & \textbf{EC}  \\ \midrule
A-Aug                 & 66.19          & 23.08          & 23.90                & 81.10          & 62.20          & 37.78              & 40.91             & 47.88   & 34.29 & 23.60 & 72.08       \\
\ +RefAug-\textit{front}                 & 72.78          & 27.34          & 28.30                & \textbf{84.62}          & 70.30          & 47.23              & 56.82             & 55.34 & 30.96 & 20.64 & 68.29          \\
\ +RefAug    & \textbf{72.93} & \textbf{29.40} & \textbf{31.20}       & 84.41 & \textbf{71.50} & \textbf{47.74}     & \textbf{60.45}    & \textbf{56.80} & \textbf{44.92} & \textbf{36.19} & \textbf{80.20} \\ \bottomrule
\end{tabular}}}
\vspace{-0.1cm}
\caption{Comparison between RefAug and prepending the reflective section to the answer (RefAug-\textit{front}).}
\vspace{-0.1cm}
\label{tab:reflect_at_front}
\end{table*}

\subsection{Analysis}

In this section, we dive deeper into additional aspects of RefAug. Results are tested on Mistral.

\subsubsection{Ablation Study}

To further assess the efficacy of the reflective section, we conduct an ablation study on its two components: alternative and follow-up reasoning. According to Table~\ref{tab:ablation}, incorporating any single reflective component to the original data significantly enhances model performance by an average of +4.5 points. This suggests that the original solutions lack sufficient information for the model to fully grasp the math reasoning skills, which is consistent with the findings of~\citet{enrichmath}. Combining both reflective components further enhances the model's comprehension of associated concepts and methodologies, improving the performance by +2.3 points over using any single one.


\subsubsection{The Amount of RefAug Data}

We explore the impact of varying the quantity of reflection-augmented instances in the whole training set. As depicted by Figure~\ref{fig:data_scaling}, the model's overall performance continually improves as more instances are augmented with reflective sections. When the model is trained through reflecting on all instances, the model maximizes its grasp of the training data and reaches the best performance, underscoring the scalability of RefAug’s benefits.


\begin{table*}[t]
\centering
\setlength{\tabcolsep}{1.3mm}{
\resizebox{0.92\textwidth}{!}{
\begin{tabular}{l|cc|ccccc|c}
\toprule
\textbf{Data}          & \textbf{GSM}   & \textbf{MATH}  & \textbf{Mathematics} & \textbf{MAWPS} & \textbf{SVAMP} & \textbf{MMLU-Math} & \textbf{SAT-Math} & \textbf{Avg.}  \\ \midrule
Standard                 & 56.25          & 13.96          & 14.80                & 73.07          & 53.50          & 37.68              & 31.82             & 40.15          \\
\quad+ RefAug \#1    & 60.05    & 17.36   & 19.40 & 80.25 & 59.30 & 43.63 & 48.64 & 46.95               \\
\quad+ RefAug \#2  & 62.70          & 17.26          & 19.20                & 82.16          & 60.40          & 42.51              & 44.55             & 46.97          \\
\quad+ RefAug \#3    & 60.80 & 16.86 & 18.60       & 80.29 & 59.70 & 42.92     & 45.45    & 46.37 \\ 
\quad+ RefAug (Avg.)    & 61.18\textsubscript{$\pm$1.1} & 17.16\textsubscript{$\pm$0.2} & 19.07\textsubscript{$\pm$0.3}       & 80.90\textsubscript{$\pm$0.9} & 59.80\textsubscript{$\pm$0.4} & 43.02\textsubscript{$\pm$0.5}     & 46.21\textsubscript{$\pm$1.7}    & 46.76\textsubscript{$\pm$0.3} \\ \bottomrule
\end{tabular}}}
\caption{We sample the reflective sections three times using the same annotation prompt in Figure~\ref{fig:refaug_prompt}, and train a separate Mistral model using each batch of the augmented data (labeled as \#1\textasciitilde\#3). The last row lists the average scores of three runs as well as their standard deviation.}
\label{tab:prompt-stability}
\vspace{-0.3cm}
\end{table*}

\subsubsection{RefAug \textit{vs.} Chain-of-Thought}

For a deeper understanding of the reflective section, we experiment with positioning it before the original solution, \textit{i.e.}, modeling $\mathcal{P}([r;a]|q)$. This arrangement can be regarded as augmenting the chain-of-thought (CoT, \citealp{chain-of-thought}) for solving the original problem. According to Table~\ref{tab:reflect_at_front}, since the reflective section contains relevant reasoning steps to the original problem, integrating it into CoT yields similar improvements as RefAug on single-round QA. However, such setup hurts performance in reflective math reasoning, which supports the original design of RefAug in developing reflective reasoning skills and reaffirms that \textbf{reflective reasoning demands distinct capabilities from standard forward reasoning}. Besides, augmenting CoT increases the token count required for predicting the final answer, thereby reducing inference efficiency (see Appendix~\ref{app:efficiency} for details). 


\subsubsection{Error Analysis}
\label{sec:error_analysis}

\begin{table}[t]
\centering
\resizebox{0.37\textwidth}{!}{
\begin{tabular}{cccc}
\toprule
\textbf{Training} & \textbf{Reasoning}        & \textbf{Calculation}        & \textbf{Total} \\ \midrule
Standard & 424       & 287       & 577   \\
RefAug   & 374{\color{gray}(-50)} & 264{\color{gray}(-23)} & 527   \\ \bottomrule
\end{tabular}}
\vspace{-0.15cm}
\caption{Error analysis on GSM8k test set. The reduction of errors is denoted in {\color{gray}gray parentheses}.}
\label{tab:error-analysis}
\vspace{-0.4cm}
\end{table}

We analyze how the model's math capabilities has been enhanced through the lens of an error analysis. Following~\citet{Xwin}, we classify errors in GSM8k into \textit{calculation} errors and \textit{reasoning} errors. Calculation errors include incorrect identification of arithmetic relationships or wrong numerical computations. Reasoning errors include mistakes pertaining to the reasoning logic, \textit{e.g.}, incoherent reasoning steps, misunderstandings of the problem, etc. Using the gold reasoning paths from GSM8k test data as a benchmark, we employ GPT-4 to determine whether solutions contain calculation errors, reasoning errors, or both. As shown in Table~\ref{tab:error-analysis}, \textbf{the improvement mostly comes from the reduction of reasoning errors}. This supports the hypothesis that training with reflection enhances the model’s problem-solving accuracy by deepening its grasp of underlying math reasoning skills.

\subsubsection{Stability of RefAug Data Annotation}
\label{sec:data_annotation_robustness}

To verify the stability of the improvements and to avoid bias from cherry-picking augmented data, we sampled reflective sections three times using GPT-4-\textit{turbo} with the same prompt in Figure~\ref{fig:refaug_prompt}. Each batch of augmented data is used to train a separate model. As shown in Table~\ref{tab:prompt-stability}, the performance gains are consistent across all augmentation samples, with a minimal standard deviation of 0.3 in overall accuracy. These results confirm that reflective practices aid in model learning and that the \textbf{observed improvements are not due to the variability of data sampling}.

\begin{table*}[t]
\centering
\setlength{\tabcolsep}{1.3mm}{
\resizebox{\textwidth}{!}{
\begin{tabular}{l|cc|ccccc|c|ccc}
\toprule
\textbf{Data}          & \textbf{GSM}   & \textbf{MATH}  & \textbf{Mathematics} & \textbf{MAWPS} & \textbf{SVAMP} & \textbf{MMLU} & \textbf{SAT} & \textbf{Avg.} & \textbf{FQA}-\textit{2nd} & \textbf{FQA}-\textit{3rd} & \textbf{EC}  \\ \midrule
Standard                 & 56.25          & 13.96          & 14.80                & 73.07          & 53.50          & 37.68              & 31.82             & 40.15  & 25.72 & 15.25 & 50.68        \\
\quad+ RefAug (GPT) & 60.05    & \textbf{17.36}   & \textbf{19.40} & 80.25 & 59.30 & \textbf{43.63} & \textbf{48.64} & \textbf{46.95} & \textbf{35.36} & \textbf{27.54} & \textbf{72.99}              \\
\quad+ RefAug (LLaMA) & \textbf{62.02} & 17.00 & 17.80 & \textbf{80.29} & \textbf{61.60} & 39.43 & 44.55 & 46.10 & 32.63 & 23.90 & 50.00 \\
\bottomrule
\end{tabular}}}
\caption{Training Mistral-7B with data where reflection sections are annotated by GPT-4-\textit{turbo} or LLaMA-3-70B-Instruct. Data annotated by LLaMA-3 yields similar improvements in standard math reasoning tasks, but fails to match GPT-annotated data in enhancing Mistral's reflective reasoning capabilities.}
\vspace{-0.2cm}
\label{tab:open-source-annotation}
\end{table*}

\subsubsection{Data Annotation with Open-Source Models}
\label{sec:open-source-annotation}

Besides using GPT to annotate RefAug data, we explore whether state-of-the-art open-source models can also serve as data annotators. We employ LLaMA-3-70B-Instruct~\cite{llama-3} for data annotation using the same prompt shown in Figure~\ref{fig:refaug_prompt}, and train a Mistral-7B model based on this data. According to results in Table~\ref{tab:open-source-annotation}, RefAug data annotated by LLaMA-3 yields a similar improvement in Mistral's performance on standard math reasoning tasks. However, the reflective reasoning capability of the resulting model falls short of its counterpart trained with GPT-annotated data. This suggests that \textbf{developing models with advanced reflective math reasoning skills demands higher quality data}, compared to what is typically required for standard forward reasoning in single-round QA.

\subsubsection{Data Contamination Analysis}
\label{sec:contamination}

To prevent the augmented data from contaminating the test sets, we check the $n$-gram overlap between the augmented reflective sections and the gold solutions within the test sets of GSM8k and MATH. Following a common approach~\cite{KPMath,MMIQC}, we utilize the test script provided by~\citet{llemma} and conduct a 20-gram check for questions and a 30-gram check for solutions. According to the results in Table~\ref{tab:contamination}, RefAug does not contaminate any test instances in GSM8k. In the MATH dataset, there is a pre-existing contamination issue: 228 questions and 167 solutions in the test set are already contaminated by the original training set. On the other hand, our RefAug data overlaps with only 5 instances in the test set, and these 5 instances were already contaminated by the training set. In other words, RefAug does not introduce new contamination to both test sets. In summary, \textbf{there is minimal contamination risk associated with RefAug in our experiments}.

In addition to the above perspectives, further analyses of RefAug's impact on model efficiency are presented in Appendix~\ref{app:efficiency}.

\begin{table}[t]
\centering
\resizebox{0.48\textwidth}{!}{
\begin{tabular}{c|ccc}
\toprule
\textbf{Dataset}               & \textbf{Source}  & \textbf{Target} & \textbf{Overlap} \\ \midrule
\multirow{3}{*}{GSM8k}  & Train Question & Test Question & 1       \\
                      & Train Answer & Test Answer & 0       \\
                      & RefAug  & Test Answer & 0       \\ \midrule
\multirow{3}{*}{MATH} & Train Question & Test Question & 228     \\
                      & Train Answer & Test Answer & 167     \\
                      & RefAug  & Test Answer & 5*       \\ \bottomrule
\end{tabular}}
\caption{The contamination check on GSM8k and MATH: the number of instances from the test set (target) sharing $n$-gram overlaps with the training data (source). We use $n=20$ for questions and $n=30$ for answers. * The 5 test instances that overlap with the augmented reflective sections were already contaminated by the original MATH training set.}
\vspace{-0.2cm}
\label{tab:contamination}
\end{table}

\section{Conclusion}
This paper proposed reflective augmentation (RefAug) for math reasoning, a method that incorporates reflection into training problems and is complementary to existing data augmentation approaches. We proved the efficacy of RefAug in not only enhancing LMs' basic problem-solving skills on single-round math problems but also in cultivating their capabilities to solve more complex reflective reasoning tasks. We further verified the effectiveness of RefAug in code generation tasks and its scalability, along with ablation studies and analyses of the methodological choices, such as the impact of data sequencing and the stability of the annotation process.

\section*{Limitations}
Some previous data augmentation studies in math reasoning created millions of data instances with OpenAI's GPT models~\cite{Xwin,mathscale,KPMath}. While testing our method at a similar scale would be valuable, budget constraints limit our ability to do so. For instance, our augmentation data for MetaMath is capped at 40K instances. In \S\ref{sec:open-source-annotation}, we note that LLaMA-3-70B shows some promising performance in annotating RefAug data for math reasoning tasks, though its capabilities have not fully matched those of GPT-4 yet. We anticipate that the development of stronger open-source models will reduce researchers' dependence on paid services of proprietary models.






\section*{Acknowledgements}

We would like to thank Hongming Zhang (Tencent AI Lab) for his valuable suggestions on experimental design and paper writing. We also thank Fangkai Jiao (Nanyang Technological University) and Zhenyu Wu (Xi’an Jiaotong University) for their suggestions that help shape our idea.

\clearpage
\bibliography{reference}

\begin{table*}[t]
\centering
\setlength{\tabcolsep}{1.3mm}{
\resizebox{0.92\textwidth}{!}{
\begin{tabular}{l|cc|ccccc|c}
\toprule
\textbf{Data}           &  \textbf{GSM}    &  \textbf{MATH}   &  \textbf{Mathematics}  &  \textbf{MAWPS}  &  \textbf{SVAMP}  &  \textbf{MMLU-Math}  &  \textbf{SAT-Math}  &  \textbf{Avg.}  \\ \midrule
Standard                  &  64.59          &  19.86  &  20.20  &  81.35  &  66.00  &  45.59  &  47.73  &  49.33         \\
\quad+ \textbf{RefAug}  &  \textbf{67.10}  &  \textbf{22.08} &  \textbf{25.60}  &  \textbf{83.64}  &  \textbf{69.40}  &  \textbf{48.97}  &  \textbf{55.00}  &  \textbf{53.11} \\ \midrule
GPT-Written Solutions  & 71.72 & 28.04 & 32.90 & 85.26 & 73.20 & 47.84 & 55.00 & 56.28 \\
\quad+ \textbf{RefAug} & \textbf{75.74} & \textbf{31.64} & \textbf{32.00} & \textbf{87.38} & \textbf{75.80} & \textbf{51.75} & \textbf{69.09} & \textbf{60.49}
\\\bottomrule
\end{tabular}}}
\caption{Results on LLaMA-3-8B. We test integrating RefAug with (1) the original training data, and (2) the data where answers are re-written by GPT-4-\textit{turbo} (see Appendix~\ref{app:gpt-answer} for GPT answer re-writing).}
\label{tab:llama3}
\end{table*}
\begin{table*}[t]
\centering
\setlength{\tabcolsep}{1.3mm}{
\resizebox{0.92\textwidth}{!}{
\begin{tabular}{l|cc|ccccc|c}
\toprule
\textbf{Data}          & \textbf{GSM}   & \textbf{MATH}  & \textbf{Mathematics} & \textbf{MAWPS} & \textbf{SVAMP} & \textbf{MMLU-Math} & \textbf{SAT-Math} & \textbf{Avg.}  \\ \midrule
Original Solutions                 & 56.25          & 13.96          & 14.80                & 73.07          & 53.50          & 37.68              & 31.82             & 40.15          \\
GPT-4-\textit{turbo} Solutions    & 65.73          & 23.10          & 23.90                & 81.14          & 68.80          & 40.25              & 41.36             & 49.18          \\
\quad+ \textbf{RefAug}    & \textbf{71.80} & \textbf{26.12} & \textbf{29.50}       & \textbf{82.84} & \textbf{70.80} & \textbf{44.76}     & \textbf{57.73}    & \textbf{54.79} \\ \bottomrule
\end{tabular}}}
\caption{Comparison between using synthetic solutions written by GPT-4-\textit{turbo} and using the originally annotated ones in GSM8k and MATH training sets, as well as applying RefAug on the synthetic solutions. Solutions written by GPT-4-\textit{turbo} are of much higher quality than the original ones.}
\label{tab:gpt-answer}
\vspace{-0.2cm}
\end{table*}

\newpage

\appendix

\section{Additional Experiments}
\label{app:additional_experiments}

In this section, we present more experimental results in addition to those in \S\ref{sec:experiments}.


\subsection{Results on LLaMA-3}
\label{app:llama3}

In addition to training Mistral-7B and Gemma-7B with RefAug, we also test LLaMA-3-8B~\cite{llama-3} on the RefAug data. According to the results in Table~\ref{tab:llama3},\textbf{ RefAug enhances the math reasoning capabilities of LLaMA-3 as well}, no matter if integrating with the original solutions or with solutions re-written by GPT-4-\textit{turbo}. This again shows the generalizability of the RefAug method, which leads to consistent improvements across various base models.

\subsection{Gemma on Reflective Math Reasoning}

Besides evaluating Mistral-based models on reflective reasoning tasks (shown in Table~\ref{tab:math-reflect}, we report scores on our Gemma-based models as well. As shown in Table~\ref{tab:math-reflect-gemma}, \textbf{the performance trends for Gemma models align with those observed on Mistral models}. RefAug demonstrates a clear advantage over traditional augmentation methods in enhancing reflective math reasoning capabilities of LMs. For instance, RefAug outscores both Q-Aug and A-Aug in the third round of follow-up QA and in the accuracy of error correction. Furthermore, as shown in Table~\ref{tab:mathchat}, a combination of A-Aug and RefAug data results in the best-performing model on the reflective reasoning scenarios of MathChat, outperforming many open-source models that are trained on substantially larger math datasets.

\begin{table}[t]
\centering

    \resizebox{0.48\textwidth}{!}{
\begin{tabular}{l|ccc|c}
\toprule
\multirow{2}{*}{\textbf{Training Data}} & \multicolumn{3}{c|}{\textbf{MathChat-FQA}}               & \multirow{2}{*}{\textbf{MathChat-EC}} \\
                      & \textit{1st}            & \textit{2nd}            & \textit{3rd}            &                         \\ \midrule
Standard                  &    60.05    &    30.05    &    20.56    &    61.99                   \\
Standard + \textbf{RefAug}             &    \textbf{64.59}    &    \textbf{40.44 }   &    \textbf{33.16}    &    \textbf{77.47}          \\ \midrule
Q-Aug                     &    61.11    &    34.67    &    26.25    &    67.68                   \\
Q-Aug$\times$2              &    63.68    &    34.45    &    26.40    &    70.41                   \\
Q-Aug + \textbf{RefAug}                &    \textbf{68.61}    &    \textbf{42.64}    &    \textbf{34.22}    &    \textbf{79.97}          \\ \midrule
A-Aug                     &    68.31    &    41.05    &    29.59    &    73.98                   \\
A-Aug$\times$2              &    70.66    &    42.79    &    32.25    &    77.39                   \\
A-Aug + \textbf{RefAug}                &    \textbf{74.15}    &    \textbf{47.80}    &    \textbf{38.54}    &    \textbf{81.11}                    \\ \bottomrule
\end{tabular}}
\caption{Results of Gemma on reflective math reasoning tasks. The general trend is similar to that of Mistral (Table~\ref{tab:math-reflect}).}
\label{tab:math-reflect-gemma}
\end{table}
\subsection{Quality of GPT-Written Answers}
\label{app:gpt-answer}

In Table~\ref{tab:math-main}, we find that answer augmentation significantly enhances performance. It improves the overall accuracy by +9.1 over the use of original training data, when averaged across Mistral and Gemma models. This surpasses the improvement of +7.2 on average seen with RefAug over the original data. A deeper analysis reveals that \textbf{the reasoning paths generated by GPT-4-\textit{turbo} are of significantly higher quality than those originally provided in the GSM8k and MATH datasets}. As demonstrated in Table~\ref{tab:gpt-answer}, merely replacing the original solutions with those generated by GPT-4-\textit{turbo} increased the accuracy from 40.15 to 49.18 on Mistral. However, RefAug does not receive such benefits as it does not alter the original reasoning paths during augmentation. Given the complementary nature of these two augmentation methods, their combination further improves the model accuracy to 54.79. This echoes the synergistic performance advantage achieved by A-Aug+RefAug over both A-Aug and A-Aug$\times$2 in Table~\ref{tab:math-main}.

\begin{table}[t]
\centering
\setlength{\tabcolsep}{1.3mm}{
\resizebox{0.3\textwidth}{!}{
\begin{tabular}{ccc}
\toprule
\textbf{Dataset} & \textbf{Alternative} & \textbf{Follow-up} \\ \midrule
GSM8K            & 96\%                            & 96\%                          \\
MATH             & 76\%                            & 72\%                          \\ \bottomrule
\end{tabular}}}
\caption{The percentage of error-free RefAug annotations by GPT-4-\textit{turbo}, including the alternative reasoning section and the follow-up reasoning section.}
\label{tab:annotation_quality}
\end{table}
\begin{table}[t]
\centering
\resizebox{0.3\textwidth}{!}{
\begin{tabular}{ccc}
\toprule
\textbf{Training} & \textbf{Data} & \textbf{Time} \\ \midrule
Standard          & 15K           & 60 min        \\
Q-Aug / A-Aug     & 30K           & 123 min       \\ 
RefAug            & 15K           & 90 min        \\\bottomrule
\end{tabular}}
\caption{The impact of various augmentation methods on dataset size and training time. These stats are tested on 8$\times$A100 GPUs.}
\label{tab:train_efficiency}
\end{table}
\begin{table}[t]
\centering
\resizebox{0.42\textwidth}{!}{
\begin{tabular}{ccc}
\toprule
\textbf{Training} & \textbf{Train Tokens} & \textbf{Test Tokens} \\ \midrule
Standard          & 171.4                 & 185.5                \\
GPT Solutions     & 358.3                 & 423.5                \\
RefAug-\textit{front}      & 910.1                 & 980.5                \\
RefAug            & 892.3                 & 219.1                \\ \bottomrule
\end{tabular}}
\caption{The resulting sequence lengths of each augmentation method during training and testing.}
\label{tab:test_efficiency}
\end{table}

\subsection{Quality of GPT-annotated Reflective Sections}
\label{app:annotation_quality}

We analyze the correctness of GPT-annotated reflective sections by manually reviewing 50 samples (25 from GSM8K, 25 from MATH) in the training set. The results, as shown in Table~\ref{tab:annotation_quality}, indicate that generating reflective sections is generally easier for GPT than solving entirely new problems. This is due to the fact that we provide both the original problem and solution during RefAug annotation. Consequently, the correctness of the annotated reflective sections is generally satisfactory.

Verification of LM-generated data is a common challenge in data augmentation. We did not dive deep into answer verification in this paper for two reasons: (1) Common methods like self-consistency voting or LM-based validation are orthogonal to our study’s focus on different augmentation types. (2) Studies have indicated that data verification often does not lead to significant performance gains, and noisy answers could help training as well~\cite{metamath,mathscale,Xwin}. This is because such answers often include many correct reasoning steps before making an error, and filtering them trades data diversity for correctness.

\subsection{Training and Inference Efficiency}
\label{app:efficiency}

For a deeper understanding of RefAug, we analyze its impact on the efficiency of model training and inference. To begin with, according to Table~\ref{tab:train_efficiency}, while RefAug does introduce additional time overhead during model training, this increase is less significant than that caused by Q-Aug or A-Aug which doubles the optimization steps due to dataset expansion. Additionally, although RefAug results in longer sequence lengths in training instances, it does not impair inference efficiency, as shown by the average number of tokens generated in Table~\ref{tab:test_efficiency}. This is due to the early stopping feature that eliminates the need to generate reflective sections during inference. Overall, \textbf{the efficiency impact brought by RefAug is minimal}.

\section{Detailed Task Settings}

In this section, we detail the datasets, training hyper-parameters, and evaluation settings of each task used in our experiments. We list the size of all datasets in Table~\ref{tab:data_stats}.

\subsection{Standard Math Reasoning}
\label{app:standard_math_task_settings}

\paragraph{Datasets} In standard math reasoning, we follow a common approach~\cite{mathcoder,metamath,Xwin} to adopt the training data from GSM8k~\cite{gsm8k} and MATH~\cite{MATH} as they are paired with human-labeled reasoning paths. For evaluation, we employ a comprehensive suite of benchmarks that span a wide range of mathematical topics. Specifically, GSM8k, SVAMP~\cite{svamp}, and MAWPS~\cite{MAWPS} focus mainly on arithmetic math word problems, while datasets such as MATH, Mathematics~\cite{mathematics}, MMLU~\cite{MMLU}, and SAT~\cite{AGIEval} encompass a broader scope including algebra, geometry, number theory, probability, and formal logic. By difficulty levels, they cover elementary (MAWPS, SVAMP), middle school (GSM8K, SAT), and more advanced levels (Mathematics, MATH, MMLU), providing an exhaustive assessment of the mathematical capabilities of language models.

\paragraph{Training Settings} During model training, we first tune the hyper-parameters using the original data under the standard fine-tuning recipe. then, these settings remain fixed across all models to avoid extensive hyper-parameter tuning for each variant. This approach is common in studies comparing models fine-tuned on varied datasets \cite{RFT,mugglemath,Lema}. Specifically, we train models for 3 epochs with a batch size of 128. The learning rate starts at 1e-5, including a warmup for the initial 3\% of steps, and then linearly decreases to 20\% of its initial value by the end of training. Training sequences are truncated to 4096 tokens. To speed up training, our model utilize \texttt{bfloat16} precision and are supported by FlashAttention-2~\cite{flashattention2}, DeepSpeed~\cite{deepspeed}, and ZeRO-3 optimization~\cite{zero}. For training on the full set of MetaMath, we follow the original authors' recommendation\footnote{\url{https://huggingface.co/meta-math/MetaMath-Mistral-7B}} to lower the learning rate to 2e-6, and for continued training on the public MetaMath checkpoint, we use a reduced learning rate of 1e-6 to be more consistent with its initial fine-tuning.

\begin{table}[t]
\centering
\resizebox{0.48\textwidth}{!}{
\begin{tabular}{lcc}
\toprule
\textbf{Dataset}     & \textbf{Train} & \textbf{Test} \\ \midrule
GSM8k~\cite{gsm8k}       & 7473  & 1319 \\
MATH~\cite{MATH}        & 7500  & 5000 \\ 
Mathematics~\cite{mathematics} & -     & 1000 \\
MAWPS~\cite{MAWPS}       & -     & 2354 \\
SVAMP~\cite{svamp}       & -     & 1000 \\
MMLU-Math~\cite{MMLU}   & -     & 974  \\
SAT-Math~\cite{AGIEval}    & -     & 220  \\ \midrule
MathChat-FQA~\cite{mathchat} & -     & 1319  \\
MathChat-EC~\cite{mathchat}    & -     & 1319  \\
MINT-Math~\cite{mint}    & -     & 273  \\ \midrule
Magicoder~\cite{magicoder}    & 38284     & -  \\
HumanEval~\cite{humaneval}    & -     & 164  \\
MBPP~\cite{MBPP}    & -     & 399  \\
\bottomrule
\end{tabular}}
\caption{Statistics of all datasets used in our training and evaluation.}
\label{tab:data_stats}
\end{table}

\paragraph{Evaluation} To facilitate answer extraction during evaluation, we append \texttt{The answer is XXX.} to the reasoning path of each training instance so that the final predicted answer is explicitly stated. We adopt the evaluation script from~\citet{mammoth} that first extracts the predicted answer and then checks for an exact match with the ground-truth. Exceptions are MMLU and SAT which use multiple-choice formats instead of numerical answers. Since our training data does not contain multiple-choice questions, the model may predict the content of an option rather than its letter identifier. Thus, on these datasets, we leverage GPT-3.5-\textit{turbo} to match the predicted content to the appropriate option before computing accuracy.


\subsection{Reflective Math Reasoning}
\label{app:reflective_math_task_settings}

Reflective math reasoning encompasses scenarios where models must consider previously provided answers to engage in further reasoning. However, benchmarks that adequately capture this dynamic are scarce in the existing literature. Utilizing the currently available resources, we evaluate our models on three tasks: follow-up QA, error correction, and feedback utilization.

The \textbf{follow-up QA (FQA)} task is assessed using the MathChat dataset~\cite{mathchat}. Each test instance consists of three turns of questions. The first turn uses the original GSM8k test set, and subsequent turns contain follow-up questions based on earlier turns. These follow-ups often require a deeper understanding of the problem, such as performing subsequent calculations based on previous answers or introducing new constraints to the original question. The solutions generated by the model for each turn are incorporated into the input for the next turn, creating a multi-turn interaction. The accuracy of each turn is evaluated separately.

The \textbf{error correction (EC)} task, also sourced from the MathChat dataset and derived from the GSM8k test set, pairs each question with an intentionally incorrect answer. The model is then tasked with identifying and correcting errors in the reasoning process. Accuracy is determined by comparing the model’s corrected answer to the ground truth.

For both tasks from MathChat, we follow the approach of~\citet{mathchat} to concatenate all previous turns into the instruction part of the input sequence. For example, in the third round of FQA, the model decodes $\mathcal{P}(a_3|[q_1;a_1;q_2;a_2;q_3])$; In EC, it decodes $\mathcal{P}(a|[q;a_{wrong};f])$, where $f$ is binary feedback indicating that $a_{wrong}$ is incorrect.


The \textbf{MINT}~\cite{mint} benchmark evaluates the ability of LMs to leverage natural language feedback to improve their predictions. We utilize the math subset from the original benchmark, which includes 273 carefully selected instances from four datasets: 48 from GSM8k, 100 from MATH, 76 from MMLU, and 49 from TheoremQA~\cite{theoremQA}. We adhere to the same evaluation protocols as the original paper except that we omit the code execution step as our math models are based on text reasoning. At each interaction turn, the model proposes a solution, and we collect binary feedback on answer correctness along with natural language feedback from an expert (\textit{i.e.}, GPT-4). This feedback is then provided to the model in the subsequent turn of prediction. The model have at most $k=5$ chances to propose solutions, and the accuracy of each turn is calculated independently. We also measure the improvement in accuracy ($\Delta$) from the first to the fifth turn to assess the model’s efficacy in leveraging feedback.

\subsection{Code Generation}
\label{app:code_task_settings}

\textbf{HumanEval} \cite{humaneval} and \textbf{MBPP} \cite{MBPP} are the most popular benchmarks for evaluating code generation capabilities of LMs \cite{wizardcoder,Dolphcoder}. Each test instance within these benchmarks includes a natural language prompt, based on which LMs generate a corresponding code snippet. The correctness of the code is verified using test cases. Additionally, EvalPlus \cite{evalplus} has developed enhanced versions of these benchmarks (\textbf{HumanEval+} / \textbf{MBPP+}) that include more comprehensive test cases for a more rigorous evaluation. Therefore, we utilize the evaluation suite provided by EvalPlus on these benchmarks, where MBPP is reduced to 399 instances for quality control.

For the training dataset, we use the \textbf{OSS-Instruct} dataset collected by Magicoder~\cite{magicoder}, which consists of synthetic instruction-code pairs generated from random code snippets sourced from GitHub. Since HumanEval and MBPP focus on Python code, we extracted the Python subset from OSS-Instruct to reduce annotation costs, resulting in a total of 38K training instances. Given the abstractive nature of code generation, we opt for analogy annotations in the follow-up reasoning part of RefAug.

We adhere to the training settings outlined in the Magicoder paper for our experiments. Models are trained over two epochs with a batch size of 512. The learning rate is initiated at 5e-5, with 15 warm-up steps followed by a linear decay. Greedy decoding is employed during inference.

\section{Baseline Implementation}
\label{app:baseline}

In this section, we detail our implementation of the major baseline methods that we compare with in the main paper, including question augmentation (Q-Aug), answer augmentation (A-Aug), and MetaMath augmentation.

\subsection{Question Augmentation}

\begin{figure*}
    \centering
    \includegraphics[width=\textwidth]{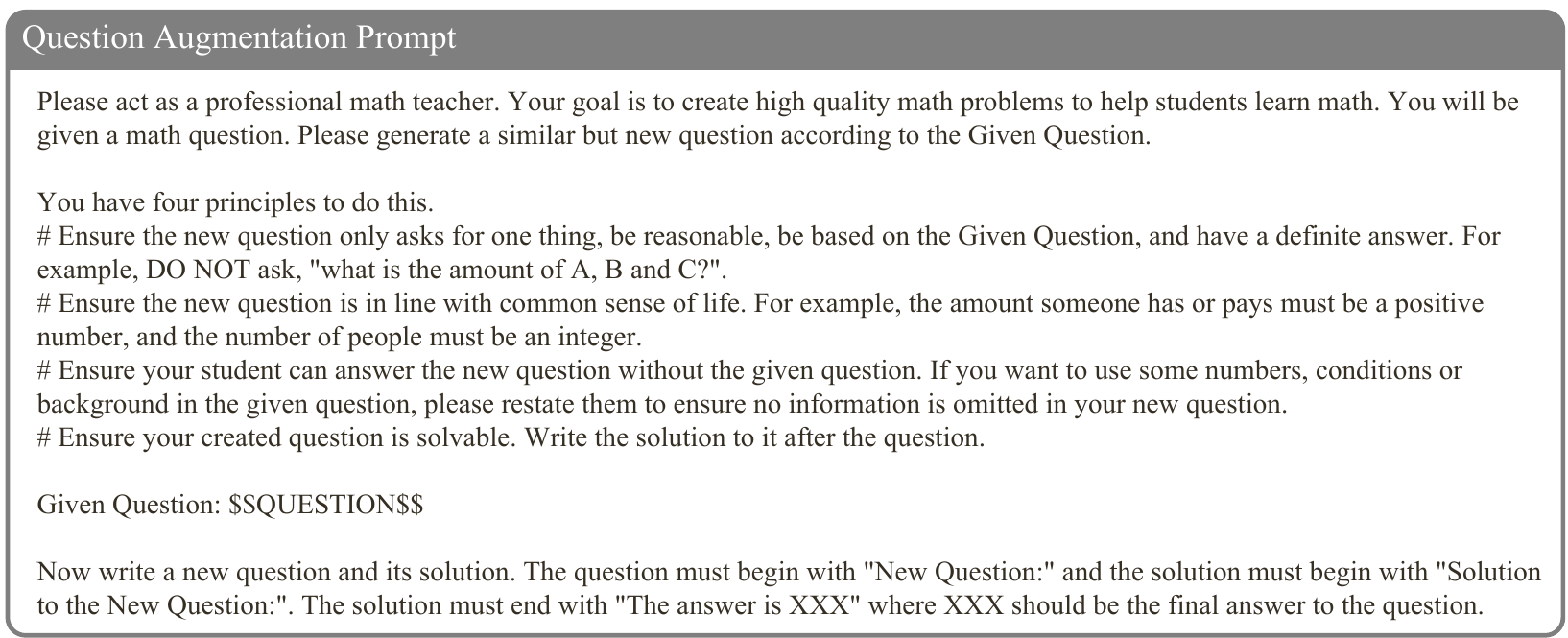}
    \vspace{-0.5cm}
    \caption{Prompt for question augmentation, adopted from~\citet{Xwin}. The only difference is that we combine question generation and solution annotation into a single prompt to save costs.}
    \label{fig:question_aug_prompt}
\end{figure*}

A single round of Q-Aug enerates a new question from each existing question in the training set, effectively doubling the dataset (illustrated in Figure~\ref{fig:intro_illustration}b). Both the augmented question and its solution are annotated by GPT-4-\textit{turbo}. During the annotation, we employ a temperature of 0.7 and a top\_p of 1.0 to ensure the diversity of math reasoning paths for both Q-Aug and A-Aug. we largely follow the question generation prompt from~\citet{Xwin} with minor adjustments. The detailed annotation prompt is provided in Figure~\ref{fig:question_aug_prompt}.

\subsection{Answer Augmentation}

\begin{figure*}
    \centering
    \includegraphics[width=\textwidth]{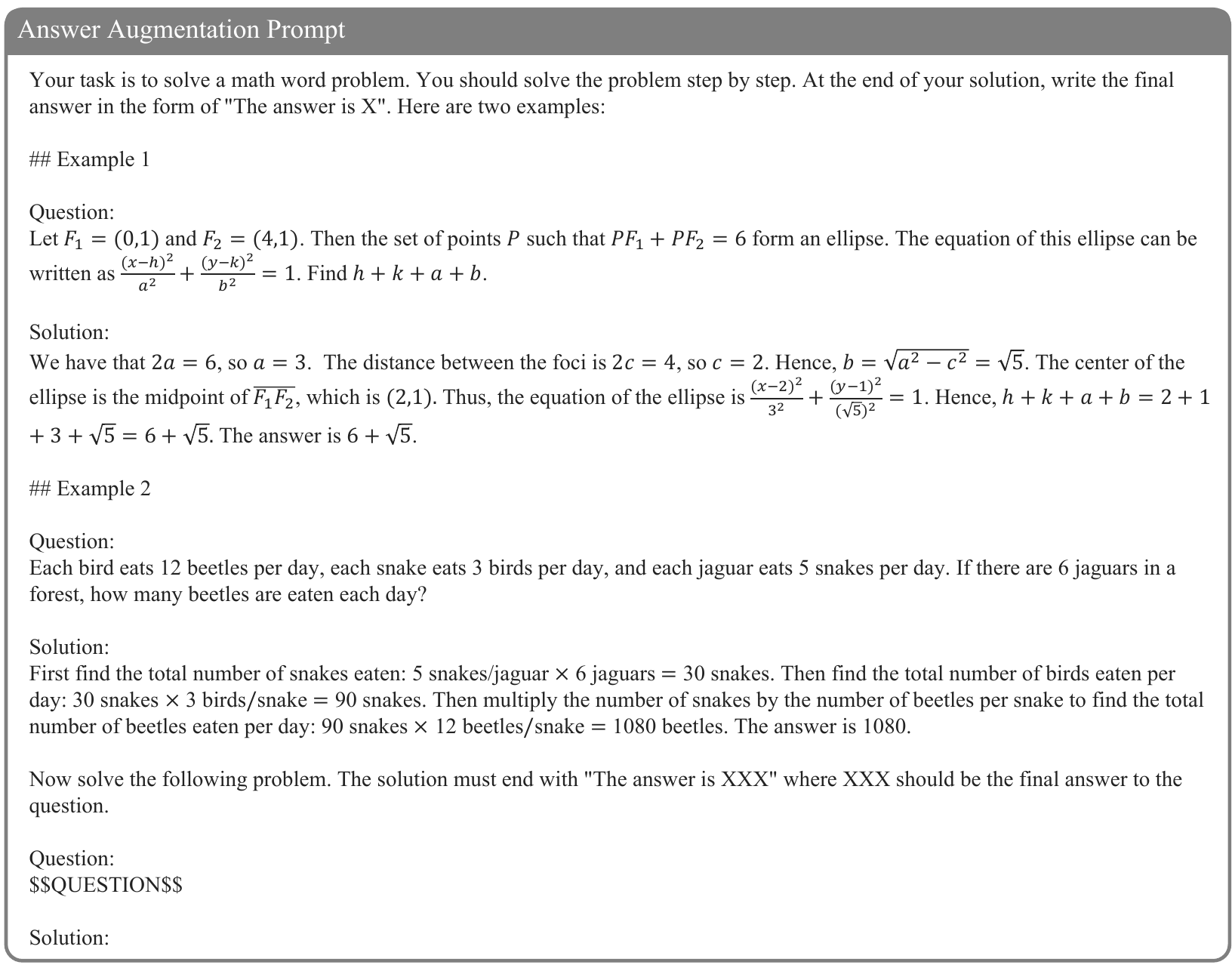}
    \vspace{-0.5cm}
    \caption{Prompt for answer augmentation, which is basically an in-context learning prompt for solving a given math problem. Two in-context examples come from MATH and GSM8k training sets, respectively.}
    \label{fig:answer_aug_prompt}
\end{figure*}

A single round of A-Aug involves re-sampling a solution for each math problem in the training set. The new solution, paired with the original question, forms a new training instance (illustrated in Figure~\ref{fig:intro_illustration}c). Consistent with other methods, the augmented solution is generated by GPT-4-\textit{turbo}. If the sampled solution diverges from the gold answer, it is discarded and re-sampled; And if a correct answer is not produced after five attempts, we retain the last sampled solution. Following the methodology described by~\citet{metamath}, the prompt for A-Aug simply instructs the model to solve an arbitrary math problem, which is detailed in Figure~\ref{fig:answer_aug_prompt}.

\subsection{MetaMath}

MetaMath~\cite{metamath} introduces a comprehensive suite of augmentation methods tailored for math reasoning tasks, which has received much attention. This suite includes answer augmentation, question rephrasing, and two backward reasoning augmentation techniques: self-verification~\cite{self-verification} and FOBAR~\cite{FOBAR}. Each method is sampled for multiple rounds to generate a large set of 400K training data. Please refer to~\citet{metamath} for more details on these methods.

When creating the \textbf{MetaMath\textsubscript{40k}} subset for our experiments in \S\ref{sec:standard_math}, we randomly select one instance from each of the four augmentation techniques for every seed math question, which we believe is the most uniform sampling strategy. For the \textbf{MetaMath\textsubscript{80k}} subset, we add one more instance from each technique for every seed question. The initially sampled 40K instances are further equipped with RefAug to be included in the full-dataset training (\textbf{MetaMath\textsubscript{400k}+RefAug\textsubscript{40k}}).

\section{Training Prompt}
\label{app:train_prompt}

\begin{figure}
    \centering
    \includegraphics[width=0.46\textwidth]{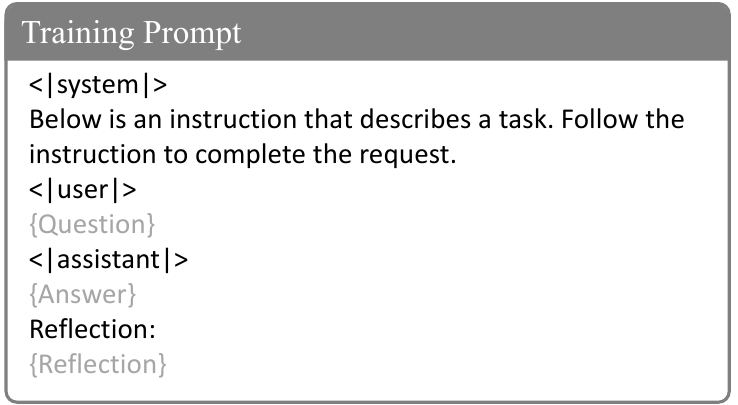}
    \caption{Prompt used for training the model. Text in {\color{gray}gray} are placeholders and will be replaced by the corresponding sections in the training instance.}
    \label{fig:train_prompt}
\end{figure}

The prompt we use to build training sequences is shown in Figure~\ref{fig:train_prompt}. The format mainly follows~\citet{tulu}, and the reflection section is appended to the original answer as the output. Loss is only calculated to tokens after \texttt{<|assistant|>}.

\section{RefAug Annotation Prompt}
\label{app:data_annotation_prompt}

\begin{figure*}
    \centering
    \includegraphics[width=\textwidth]{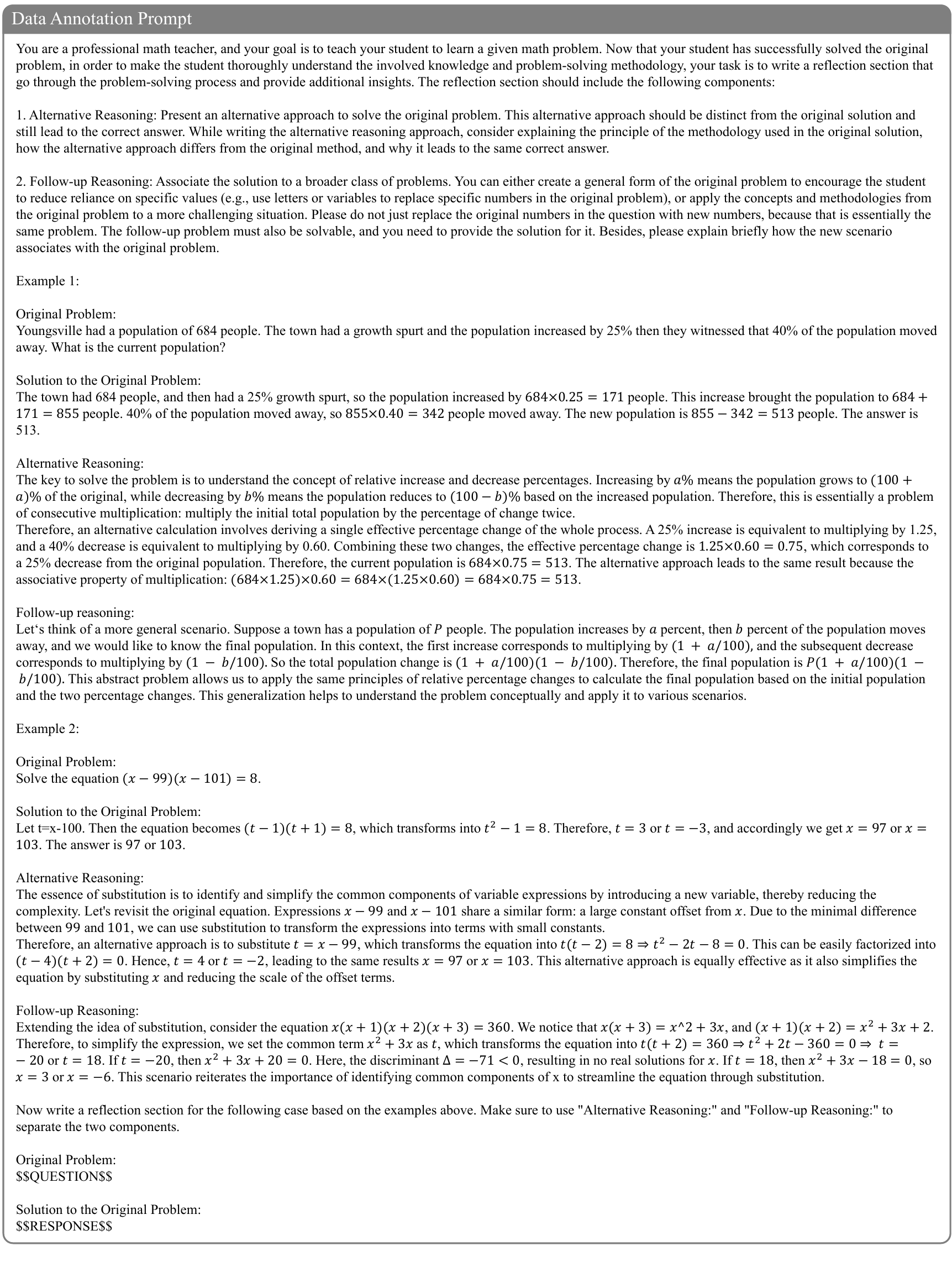}
    \vspace{-0.7cm}
    \caption{Prompt for annotating the reflective section. The prompt first explains the contents to annotate within the reflective section, and then presents two in-context examples for demonstration. GPT-4-\textit{turbo} is employed for annotation.}
    \label{fig:refaug_prompt}
\end{figure*}

The prompt we use for annotating reflective sections are detailed in Figure~\ref{fig:refaug_prompt}, which includes a description of the general principles of reflective reasoning and two in-context examples. We use temperature=0.7 and top\_p=1.0 when sampling with GPT-4-\textit{turbo}.

\section{License of Artifacts}

We note that the collection of RefAug data, if annotated by an external model, should comply with its terms of use. For example, using GPT-generated data is subject to the terms of use of OpenAI services\footnote{\url{https://openai.com/policies/terms-of-use/}}, and using LLaMA-generated data is subject to Meta’s LLaMA license agreement\footnote{\url{https://llama.meta.com/llama3/license/}}.

\end{document}